\documentclass[conference]{IEEEtran}
\IEEEoverridecommandlockouts

\usepackage{cite}
\usepackage{amsmath,amssymb,amsfonts}
\usepackage{algorithmic}
\usepackage{graphicx}
\usepackage{textcomp}
\usepackage{xcolor}
\usepackage{float}
\def\BibTeX{{\rm B\kern-.05em{\sc i\kern-.025em b}\kern-.08em
    T\kern-.1667em\lower.7ex\hbox{E}\kern-.125emX}}

\usepackage{hyperref}

\begin{document}
\title{Comparative Analysis of Military Detection Using Drone Imagery Across Multiple Visual Spectrums}

\author{\IEEEauthorblockN{Sourov Roy Shuvo}
\IEEEauthorblockA{\textit{School of Computer Engineering} \\
\textit{KIIT Deemed to be University}\\
Bhubaneswar, India \\
sourovroyshuvo777@gmail.com}
\and
\IEEEauthorblockN{Prajwal Panth}
\IEEEauthorblockA{\textit{School of Computer Engineering} \\
\textit{KIIT Deemed to be University}\\
Bhubaneswar, India \\
prajwal.panth21@gmail.com}
\and
\IEEEauthorblockN{Rajesh Chowdhury}
\IEEEauthorblockA{\textit{School of Computer Engineering} \\
\textit{KIIT Deemed to be University}\\
Bhubaneswar, India \\
rajesh99.bd@gmail.com}
\and
\IEEEauthorblockN{Sorup Chakraborty}
\IEEEauthorblockA{\textit{School of Computer Engineering} \\
\textit{KIIT Deemed to be University}\\
Bhubaneswar, India \\
sorupchakraborty001@gmail.com}
\and
\IEEEauthorblockN{Sudip Chakrabarty}
\IEEEauthorblockA{\textit{School of Computer Engineering} \\
\textit{KIIT Deemed to be University}\\
Bhubaneswar, India \\
sudipchakrabarty6@gmail.com}
\and
\IEEEauthorblockN{Prasant Kumar Pattnaik}
\IEEEauthorblockA{\textit{School of Computer Engineering} \\
\textit{KIIT Deemed to be University}\\
Bhubaneswar, India \\
patnaikprasantfcs@kiit.ac.in}
}

\maketitle

\begin{abstract}
In modern warfare, drones are becoming an essential part of intelligence gathering and carrying out precise attacks in different kinds of hostile environments. Their ability to operate in real-time and hostile environments from a safe distance makes them invaluable for surveillance and military operations. The KIIT-MiTA dataset is comprised of images of different military scenarios taken from drones, and these provide a foundation for detecting military objects, but it does not take into account the various types of real-world scenarios. With that in mind, to evaluate how the models are performing under varying conditions, four different types of datasets are created: Gray Scale, Thermal Vision, Night Vision, and ObscuraVision. These simulate the real-world environments such as low visibility, heat-based imagery, and nighttime conditions. The YOLOv11-small model is trained and used to detect objects across diverse settings. This research boosts the performance and reliability of drone-based operations by contributing to the development of advanced detection systems in both defensive and offensive missions.
\end{abstract}


\section{Introduction}
Drones have emerged as a powerful tool in modern military operations, playing a crucial role in surveillance, reconnaissance, and targeted strikes. With the ability to operate in dangerous and hard-to-reach areas, drones provide significant advantages in terms of safety and efficiency. The evolution of drone technology has also led to improvements in object detection capabilities, making it possible to identify and track military assets with greater precision. However, as drones are increasingly used in diverse environments like high-heat or low-light conditions, the reliability of object detection systems remains a challenge.

The motivation for this research is driven by the need to enhance the performance of object detection systems under various environmental conditions. While existing models have achieved success in controlled settings, their ability to detect objects accurately in difficult conditions, such as thermal and night vision, has not been extensively studied. The integration of drones in modern warfare demands detection systems that can adapt to these diverse challenges, ensuring continuous effectiveness in dynamic environments.

This paper addresses the gap in current research by evaluating object detection performance using the KIIT-MITA dataset under different visual conditions. The dataset, originally captured using standard imaging, is extended into four versions: Gray Scale, Thermal Imaging, Night Vision, and ObscuraVision. By assessing how these variations impact detection accuracy, the research aims to improve the robustness of drone-based object detection systems in real-world military operations.

The key contribution of this paper is the evaluation of the YOLOv11-small model on four versions of the KIIT-MITA dataset, which simulate different environmental conditions. These conditions reflect real-world challenges often faced in military operations, such as fog, pollution, and the common use of thermal imaging to detect heat signatures. Night Vision is crucial for detecting objects in low-light environments, while Thermal Imaging is frequently used in poor visibility. By testing the model’s performance under these conditions, this study provides valuable insights into its ability to maintain accuracy and reliability in diverse settings. The research advances the robustness of drone-based detection systems, particularly in military surveillance and defense, where environmental factors impact operational success.

The remaining part of this paper is organized as follows. Section \ref{sec2} reviews related work in drone-based object detection field. Section \ref{sec3} presents the methodology, including the dataset creation and model training process. Section IV discusses experimental setup and results and finally Section \ref{sec5} summarizes the paper and suggests potential avenues for future research.

\begin{figure*}[ht!]
    \centering    \includegraphics[width=\textwidth, height=5cm]{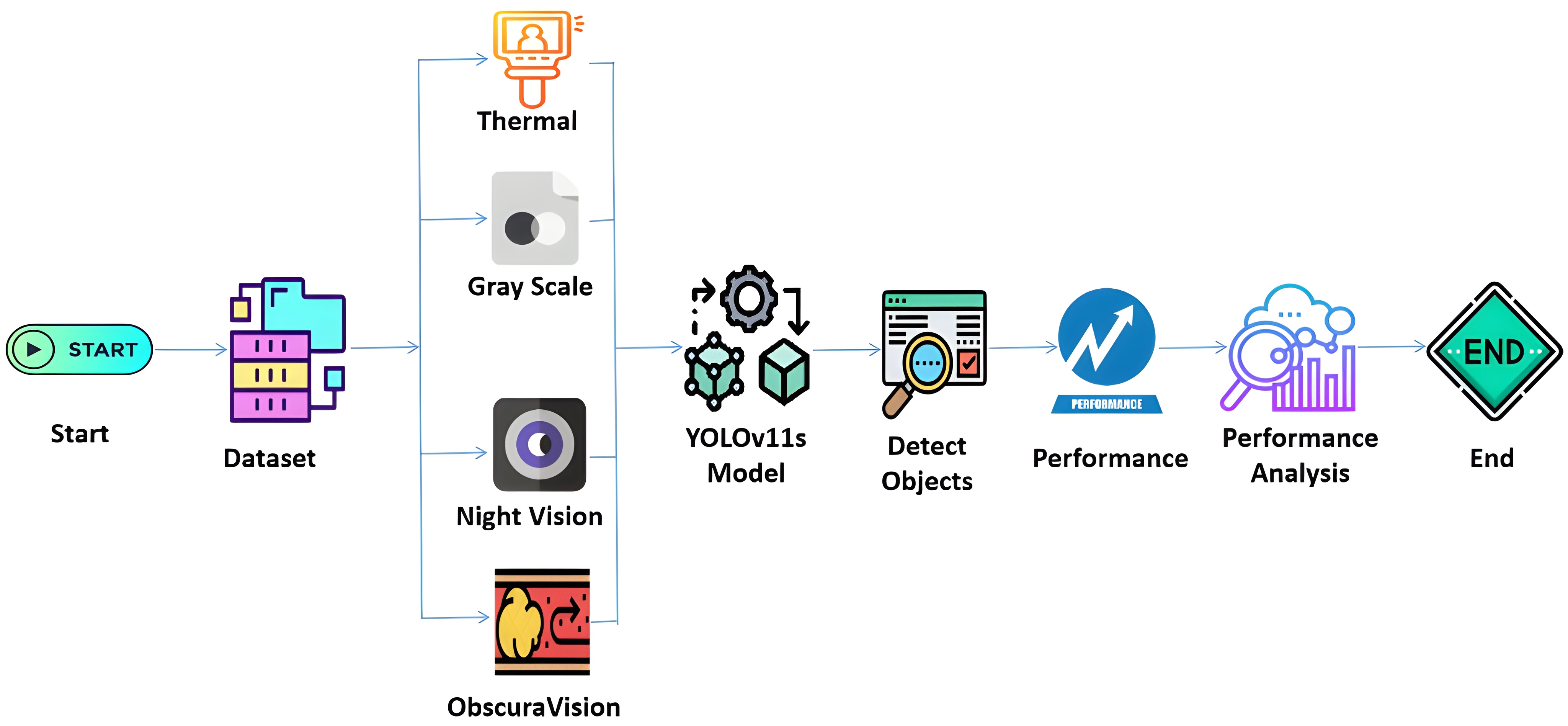} 
    \caption{System pipeline illustrating dataset input, generation of four distinct visual representations, training with YOLOv11-small, result evaluation across multiple vision types.}
    \label{fig:1}
\end{figure*}

\section{Literature Survey and Related Works}\label{sec2}
A lot of researchers have worked on object detection using deep learning, especially for improving accuracy and speed in real-world scenarios. He et al. (2024) \cite{he2024research} applied YOLOv11-Seg for segmenting 13 object types on construction sites, achieving strong accuracy. Chakrabarty et al. (2025) \cite{chakrabarty2025drones} emphasizes real-time detection and tracking of military targets using drone-captured images, showcasing model evaluation and optimization for dynamic and resource-constrained defense scenarios without delving into various spectral or illumination variations. Jiang et al. (2022) \cite{jiang2022object} used YOLO models to detect objects in UAV-based thermal infrared images and videos, achieving 88.69\% mAP and 50 FPS speed. Although effective, challenges remained due to coarse resolution and lack of public labeled datasets. Bustos et al. (2023) \cite{bustos2023systematic} conducted a systematic review of object detection methods using thermal images and near-infrared, highlighting advances across IR wavelengths. They noted challenges such as limited public datasets and high noise in IR images, which hinder robust detection. Recent studies have focused on object detection using night vision \cite{xiao2020making, bhabad2023object} imagery, employing YOLO-based models for enhanced performance in thermal and infrared \cite{li2021yolo, sun2024multi} conditions. Nayan et al. (2020) \cite{nayan2020detection} proposed an SSD-based approach for detecting objects in noisy images, showing good performance under poor lighting and degraded conditions. However, the method was limited to conventional detectors and lacked evaluation on newer deep learning models like YOLO variants. Rodríguez-Rodríguez et al. (2024) \cite{rodriguez2024impact} examined how noise and brightness variations degrade object detection performance in models like YOLO and Faster R-CNN, especially for small objects, but did not propose robust countermeasures. Collectively, these studies emphasize the growing need for object detection models that remain accurate under thermal, low-light, grayscale \cite{lorencs2009fast}, noisy, and night vision imagery, ensuring reliable performance in real-world surveillance, defense, and autonomous applications.

\section{Proposed Methodology}\label{sec3}
The system initiates with a dataset comprising four distinct image modalities: Thermal, Grayscale, Night Vision, and ObscuraVision. Each image type provides unique visual cues to enhance object detection under varied conditions. These images are processed through the YOLOv11s model, a lightweight and efficient deep learning algorithm optimized for real-time object detection. The model executes preprocessing, inference, and postprocessing to identify and localize objects within each frame. Following detection, performance metrics such as mAP@50, mAP@50-95, precision, recall, and F1-score are calculated to assess accuracy and efficiency. During the performance analysis stage, results are compared across all image types to determine which modality offers the most effective detection outcomes. Figure \ref{fig:1} presents an overview of this pipeline, outlining the transition from dataset input to performance evaluation.

\subsection{Dataset Details}
The dataset used in this research is named  \textbf{KIIT-MiTA}\footnote{KIIT-MiTA: https://kiit-mita.netlify.app/}, which contains 1,700 high-quality images captured by drones. Each image has been carefully annotated, resulting in over 4,100 detailed labels. The annotations follow the YOLO format, with individual text files for each image that include normalized coordinates and class labels. This dataset is divided into training, validation, and testing subsets to support effective learning, validation during training, and final performance of the model. 
The dataset includes the following class names:

\begin{center}
\begin{itemize}
\item[1.] \textit{Artillery}
\item[2.] \textit{Missile}
\item[3.] \textit{Radar}
\item[4.] \textit{Multiple Rocket Launcher}
\item[5.] \textit{Soldier}
\item[6.] \textit{Tank}
\item[7.] \textit{Vehicle}
\end{itemize}
\end{center}

\subsection{Dataset Preparation}
\begin{itemize}
\item \textbf{Thermal Vision : }
To simulate thermal infrared vision, images were converted to grayscale and then mapped to a pseudo-thermal color space using OpenCV’s COLORMAP INFERNO \cite{zeileis2020colorspace}. Grayscale conversion isolates intensity information, while normalization enhances contrast across the full 0–255 range. Sample images from the Thermal Vision dataset are illustrated in Figure \ref{fig:2}. The inferno colormap was chosen for its perceptual similarity to real thermal imagery, representing cooler areas in dark tones and warmer regions in bright hues. This augmented dataset, referred to as ThermalVision, mimics infrared sensor output and supports model training under simulated thermal conditions common in low-light or night-time surveillance scenarios.
\begin{figure}[ht!]
    \centering
    \begin{minipage}[]{0.49\textwidth} 
        \centering
        \includegraphics[width=\textwidth, height=4.4cm]{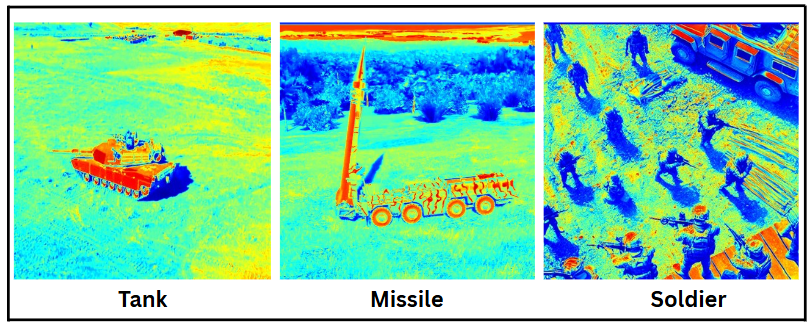}
        \caption{Simulated thermal images using Inferno colormap for surveillance tasks.}
        \label{fig:2}
    \end{minipage}\hfill
\end{figure}

\item \textbf{Night Vision : }
To simulate night vision optics, images were converted to grayscale and enhanced in brightness and contrast using OpenCV’s convertScaleAbs function \cite{singh2019advanced}. 
\begin{figure}[ht!]
    \centering
    \begin{minipage}[]{0.49\textwidth} 
        \centering
        \includegraphics[width=\textwidth, height=4.4cm]{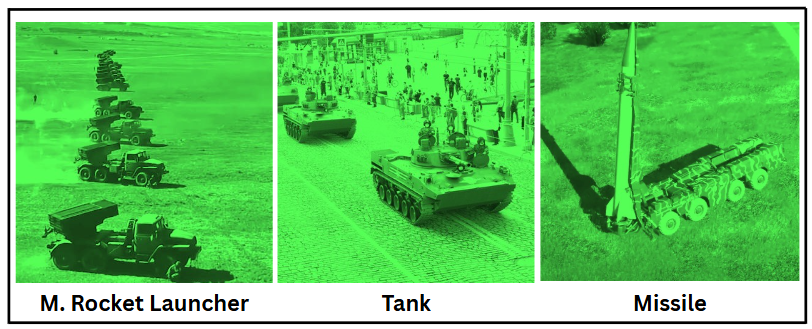}
        \caption{Sample night vision images generated by grayscale enhancement and green tint overlay to simulate low-light surveillance environments.}
        \label{fig:3}
    \end{minipage}\hfill
\end{figure}
A green tint typical of night vision devices was applied by merging grayscale intensity with a weighted green channel. This augmentation mimics the visual output of monochromatic image intensifiers, commonly used in low-light military and surveillance scenarios. The processed dataset, termed NightVision, preserves spatial structure while providing a domain shift toward nocturnal or covert observation environments, aiding in the evaluation of detection models under reduced lighting conditions. Sample outputs from the NightVision dataset are shown in Figure \ref{fig:3}.

\item \textbf{Gray Scale : }
For the Gray Scale condition, the images were converted from color to grayscale using OpenCV’s built-in function \cite{khobragade2012comparative}. This conversion reduces the image to a single intensity channel, making it easier to focus on the shapes and structures in the image for object detection tasks. Grayscale preprocessing removes color distractions, enabling models to prioritize spatial features such as edges, contours, textures, and object geometry. This approach is particularly valuable in scenarios where color information is unreliable, inconsistent, or unnecessary. Representative grayscale samples are depicted in Figure \ref{fig:4} for visual reference and clarity.

\begin{figure}[ht!]
    \centering
    \begin{minipage}[]{0.49\textwidth} 
        \centering
        \includegraphics[width=\textwidth, height=4.4cm]{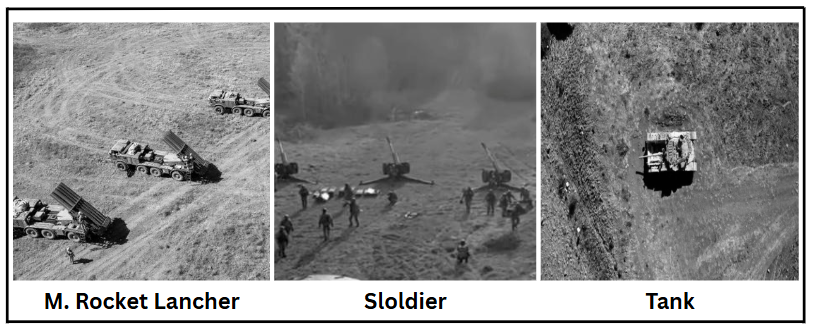}
        \caption{Grayscale-transformed samples emphasizing shape and contour details for improved detection.}
        \label{fig:4}
    \end{minipage}\hfill
\end{figure}

\item \textbf{ObscuraVision : }
To simulate real-world visibility challenges such as light fog, motion distortion, and contrast irregularities, often encountered in aerial surveillance, a specialized version of the dataset was created using controlled augmentation techniques. This modified dataset will be referred to as \textbf{ObscuraVision} throughout this paper. Sample visuals portraying the simulated conditions of ObscuraVision are presented in Figure \ref{fig:5}. It represents low-level environmental interference (25\%) that may impact detection performance.
\begin{figure}[ht!]
    \centering
    \begin{minipage}[]{0.49\textwidth} 
        \centering
        \includegraphics[width=\textwidth, height=4.4cm]{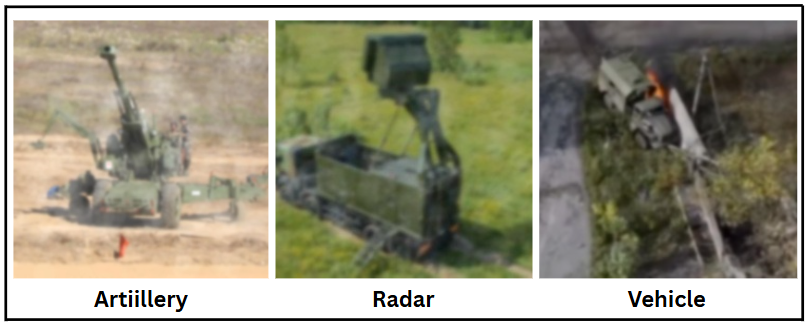}
        \caption{Sample obscuravision images simulating mild real-world visual distortions such as motion blur, fog, and contrast variations to evaluate detection robustness.}

        \label{fig:5}
    \end{minipage}\hfill
\end{figure}

The combined severity of applied transformations is defined by the following Equation \ref{eq1}:

\begin{equation}
\label{eq1}
O_s = \alpha \cdot M_b + \beta \cdot F_g + \gamma \cdot C_b
\end{equation}

\noindent
where $O_s$ is the \textit{obscuration severity score}, $M_b$ denotes \textit{motion blur}, $F_g$ is the \textit{fog coefficient}, $C_b$ represents \textit{contrast and brightness adjustment}, and $\alpha$, $\beta$, and $\gamma$ are weighting constants (set to 1.0 for uniform importance).

This formulation allows a quantifiable representation of image degradation, helping evaluate model robustness under visual distortions. The \textbf{ObscuraVision} dataset uses fixed low values, such as a blur limit of 3, fog coefficient around 0.1, and contrast/brightness limit set to 0.1, to imitate realistic conditions like early morning haze, light pollution, or minor vibration-induced blur \cite{7894491}, without heavily compromising object visibility.

\end{itemize}

\subsection{Model Selection}
For this research, we selected the \textbf{YOLOv11s} model for object detection. This model is one of the latest additions to the YOLO family and is well-known for achieving an excellent trade-off between precision and performance. YOLOv11s is specially designed to perform well even on smaller or resource-constrained systems, while still maintaining high detection performance. Enhancements in extracting features, attention modeling, and eliminating anchor reliance make it appropriate for real-time drone image analysis. Due to its lightweight structure and high efficiency, YOLOv11s was considered the most appropriate choice for this project.

\subsection{Model Training}
The training of the YOLOv11s model was conducted using Kaggle's cloud environment, which provided access to a NVIDIA T4 GPU. The model was trained for 100 iterations with a mini-batch size of 16, utilizing input images scaled to $640 \times 640$ \text{pixels}.  The training workflow incorporated image augmentation strategies,which enhanced the model's ability to generalize to previously unseen data. We implemented a learning rate scheduler that adaptively decreased the learning rate over the course of training, preventing overfitting and ensuring stable convergence. The loss function optimized during training included classification loss, objectness loss, and bounding box regression loss. The model weights were saved based on the best validation performance to ensure reliable evaluation.

\subsection{Evaluation Metrics}
To assess how well the object detection models performed, we used common evaluation metrics such as Precision, Recall, the F1 Score, mAP@50, and mAP@50–95. Precision quantifies the accuracy of positive predictions, as shown in Equation \ref{eq:precision}, while Recall measures the ability to detect all relevant instances, as shown in Equation \ref{eq:recall} and Equation \ref{eq:f1score} presents the F1 Score, which provides a balanced evaluation of the model’s performance.

\begin{equation}
\text{Precision, Pr} = \frac{TrPs}{TrPs + FlPs}
\label{eq:precision}
\end{equation}

\begin{equation}
\text{Recall, Re} = \frac{TrPs}{TrPs + FlNg}
\label{eq:recall}
\end{equation}

\begin{equation}
\text{F1 Score} = 2 \times \frac{\text{Pr} \times \text{Re}}{\text{Pr} + \text{Re}}
\label{eq:f1score}
\end{equation}

\vspace{0.3cm}
\noindent\textbf{Where,} \\
\noindent$TrPs$ = True Positives \\
\noindent$FlPs$ = False Positives \\
\noindent$FlNg$ = False Negatives

\vspace{0.2cm}
However, for this study, the primary focus is on mAP@50 and mAP@50–95, which represent the mean Average Precision at IoU thresholds of 0.5 and 0.5–0.95, respectively, as shown in Equations \ref{map50} and \ref{map90}. These metrics offer a comprehensive evaluation of detection accuracy across varying object overlap thresholds.

\begin{equation}
\text{mAP@50} = \frac{1}{N} \sum_{i=1}^{N} \text{AP}_i^{\text{ IoU}=0.50}
\label{map50}
\end{equation}

\begin{equation}
\text{mAP@50–95} = \frac{1}{T \cdot N} \sum_{t=1}^{T} \sum_{i=1}^{N} \text{AP}_i^{\text{ IoU}=t}
\label{map90}
\end{equation}

\vspace{0.3cm}
\noindent\textbf{Where,} \\
\noindent$N$ = Number of object classes \vspace{0.1cm}\\
\noindent$T$ = Number of IoU thresholds  \vspace{0.12cm}\\
\noindent$\text{AP}_i^{\text{ IoU}=t}$ = Average Precision for class $i$ at IoU threshold $t$

\subsection{Detected Object Visualization}
\begin{figure*}
        \centering
        \includegraphics[width=0.98\textwidth, height=12.4cm]{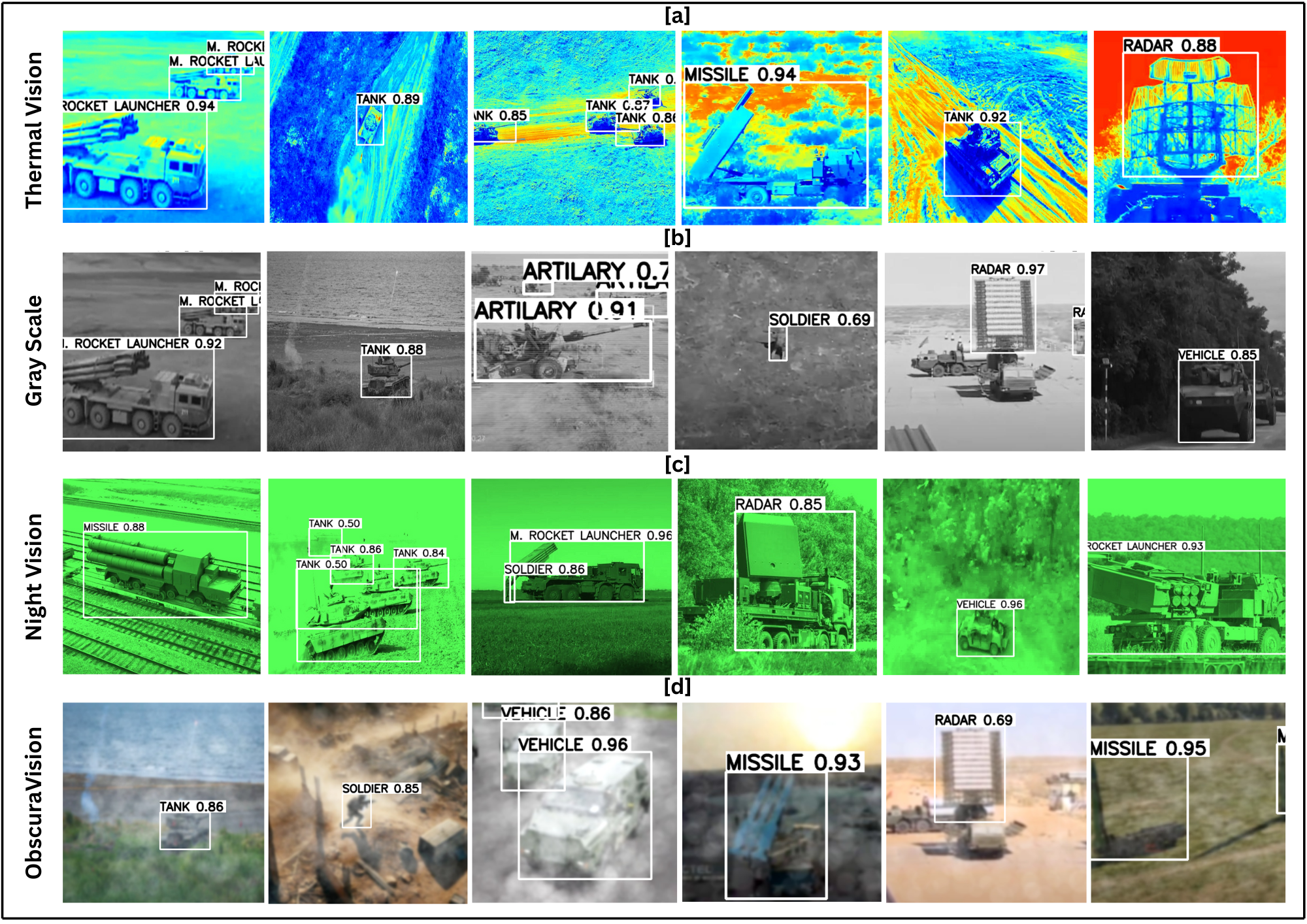} 
        \caption{Detection results from various datasets: [a] Thermal Vision dataset showcasing object detection in thermal conditions, [b] grayscale detection highlighting reduced color impact, [c] Night Vision dataset simulating low-light surveillance, and [d] ObscuraVision illustrating detection under fog and motion distortion.}
        \label{fig:6}
\end{figure*}

The object detection model was employed and trained separately on four distinct datasets corresponding to the following imaging modalities: Gray Scale, Thermal Vision, Night Vision, and ObscuraVision. These datasets consist of high-resolution static images captured under varying operational conditions. Each detected object is annotated with bounding boxes, effectively demonstrating the model’s ability to localize and distinguish military targets within visually complex and challenging environments. The diversity of the training data enables a comprehensive assessment of the model’s robustness, particularly under conditions such as low visibility, thermal distortion, and partial occlusion. The output results show consistent detection performance across most modalities, with Night Vision and Gray Scale images achieving the highest mAP scores. Thermal Vision and ObscuraVision also produced reliable results, underscoring the model’s adaptability. Representative examples of the detection outputs are presented in Fig.~\ref{fig:6}, illustrating the model’s capability to maintain accurate object classification and localization across varied environmental conditions.

\section{Results Analysis}\label{sec4}

The experimental results presented in Table~\ref{tab1} and Table~\ref{tab2} offer a comparative analysis of four imaging modalities Gray Scale, Thermal Vision, Night Vision, and ObscuraVision evaluated using standard object detection metrics. The key performance indicators emphasized in this study are mAP@50 and mAP@50-95, which serve as the primary criteria for model ranking and evaluation.

Among the modalities, Night Vision demonstrated the highest detection accuracy with an accuracy measured at mAP@50 of 0.701 and an accuracy measured at mAP@50-95 of 0.484, thus ranking highest based on the primary evaluation criteria. Although Night Vision incurred the highest total processing time (10.1 ms), its superior accuracy justifies its selection for applications where detection performance is paramount. ObscuraVision closely followed with a mAP@50 of 0.694 and mAP@50-95 of 0.467, offering a strong balance between accuracy and inference speed (9.4 ms). Thermal Vision also performed competitively with an accuracy measured at mAP@50 of 0.680 \& an accuracy measured at mAP@50-95 of 0.466, though slightly lower than ObscuraVision. Notably, Gray Scale, despite having the fastest processing time (8.6 ms), recorded the lowest detection accuracy, with a mAP@50 of 0.603 and mAP@50-95 of 0.374, positioning it last in the ranking based on the mAP metrics.

\begin{figure}[ht!]
    \centering
    \begin{minipage}[]{0.49\textwidth} 
        \centering
        \includegraphics[width=\textwidth, height=7.5cm]{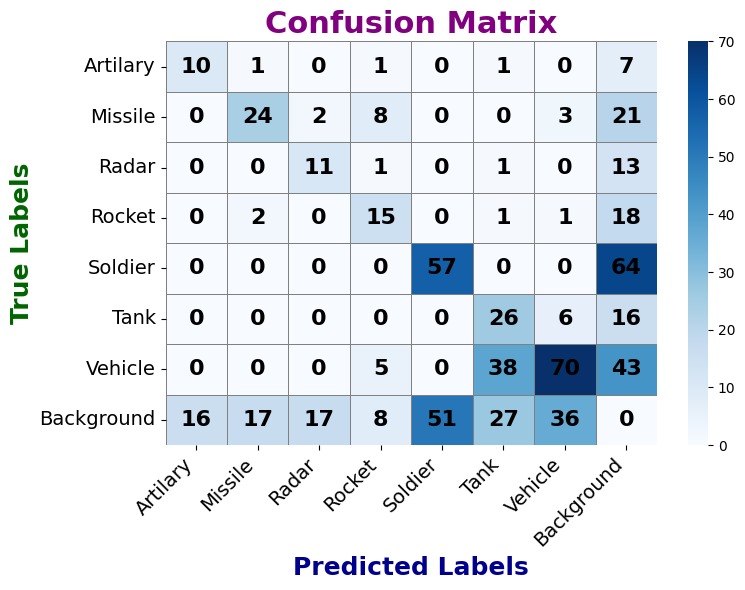}
        \caption{Confusion matrix of YOLOv11-small trained on the KIIT-MiTA (Gray Scale) dataset.}
        \vspace{-7mm}
        \label{fig:matrix}
    \end{minipage}\hfill
\end{figure}

While F1 score and related metrics (Table~\ref{tab2}) were also calculated for completeness, their influence on model selection remains secondary in this context. Gray Scale led in F1 score (0.700), and Thermal Vision achieved the highest precision (0.752); however, these metrics were not the primary basis for model evaluation. The detailed class-wise performance of the YOLOv11-small model on the KIIT-MiTA grayscale dataset is illustrated in the confusion matrix shown in Fig.~\ref{fig:matrix}. The confusion matrix helps to demonstrate how well the model distinguishes between the different classes present in the gray scale dataset.

In summary, Night Vision is the most suitable modality when prioritizing detection accuracy, while ObscuraVision and Thermal Vision offer favorable trade-offs between performance and speed. Gray Scale, although efficient, is less favorable in terms of mAP-based accuracy and is more appropriate for latency-sensitive but low-accuracy applications.

\begin{table*}[ht!]
    \centering
    \caption{COMPARATIVE PERFORMANCE AND PROCESSING TIMES OF YOLOV11S ACROSS DIFFERENT VISION MODALITIES.}
    \label{tab1}
    \resizebox{0.9\textwidth}{!}{
    \begin{tabular}{|l|c|c|c|c|c|c|}
        \hline
        \rule{0pt}{8pt} \textbf{Model} & \textbf{Preprocess (ms)} & \textbf{Inference (ms)} & \textbf{Postprocess (ms)} & \textbf{Total Time (ms)} & \textbf{\hspace{0.2cm} mAP50 \hspace{0.2cm} } & \textbf{\hspace{0.2cm}mAP50-95\hspace{0.2cm}} \\
        \hline
        \rule{0pt}{8pt} Gray Scale & 0.3 & 5.3 & 3.0 & 8.6 & 0.603 & 0.374 \\
        \hline
        \rule{0pt}{8pt} Thermal Vision & 0.6 & 5.2 & 3.9 & 9.7 & 0.680 & 0.466 \\
        \hline
        \rule{0pt}{8pt} Night Vision & 0.3 & 5.3 & 4.5 & 10.1 & 0.701 & 0.484 \\
        \hline
        \rule{0pt}{8pt} ObscuraVision & 0.6 & 5.2 & 3.6 & 9.4 & 0.694 & 0.467 \\
        \hline
    \end{tabular}}
\end{table*}

\begin{table*}[ht!]
    \centering
    \caption{KEY PERFORMANCE INDICATORS AND TRAINING OVERHEAD FOR DIFFERENT VISION INPUTS.}
    \label{tab2}
    \resizebox{0.5\textwidth}{!}{
    \begin{tabular}{|l|c|c|c|c|}
        \hline
        \rule{0pt}{8pt} \textbf{Model} & \textbf{Precision} & \textbf{Recall} & \textbf{F1 Score} & \textbf{Training Time (h)} \\
        \hline
        \rule{0pt}{8pt} Gray Scale & 0.732 & 0.671 & 0.700 & 1.091  \\
        \hline
        \rule{0pt}{8pt} Thermal Vision & 0.752 & 0.623 & 0.681 & 1.219  \\
        \hline
        \rule{0pt}{8pt} Night Vision & 0.713 & 0.662 & 0.686 & 1.042  \\
        \hline
        \rule{0pt}{8pt} ObscuraVision & 0.624 & 0.555 & 0.587 & 0.989  \\
        \hline
    \end{tabular}}
\end{table*}

\section{Conclusion}\label{sec5}

This study presents a thorough assessment of an object detection model trained on static military imagery captured by drone under four distinct visual conditions: Gray Scale, Thermal Vision, Night Vision, and Occlusion. Using key evaluation metrics including mAP@50, mAP@50–95, and F1 Score the model demonstrated effectiveness in accurately identifying and classifying military targets, even in visually degraded or challenging scenarios. Among the modalities, Night Vision achieved the highest detection accuracy, particularly with respect to mAP metrics. Thermal Vision also delivered strong performance, whereas results under occlusion highlighted opportunities for further enhancement. While the F1 Score contributed supplementary insights, model comparisons were primarily guided by mAP values, given their comprehensive reflection of detection quality. The results of the experiments demonstrate the model’s reliability and flexibility, highlighting its potential for practical applications in defense and monitoring operations.

Future research may focus on integrating multi-modal fusion techniques that combine different visual inputs to enhance detection accuracy under complex or uncertain environments. Incorporating temporal data from continuous drone footage could also enable dynamic scene understanding and motion-aware tracking. Additional efforts may include optimizing the model for real-time inference on edge devices and expanding the dataset with real-world annotated samples, particularly for low-visibility and occluded conditions. These enhancements aim to further improve model generalizability and operational readiness in field applications.

\bibliographystyle{IEEEtran}
\bibliography{references}
\end{document}